\documentclass{article}

\usepackage[final]{nips_2016}

\usepackage{times}
\usepackage{graphicx} % more modern
\usepackage{subfigure} 

\usepackage{natbib}

\usepackage{algorithm}
\usepackage{algorithmic}

\usepackage{hyperref}

\usepackage{times}
\usepackage{url}

\usepackage{latexsym}

\usepackage{xspace}
\usepackage{mathtools}
\usepackage{amsmath}
\usepackage{amssymb}
\usepackage{multirow}
\usepackage{tikz}
\usepackage{pifont}
\usepackage[inline]{enumitem}% http://ctan.org/pkg/enumitem
\usetikzlibrary{positioning}
\usetikzlibrary{calc}
\usetikzlibrary{fit}
\usetikzlibrary{decorations.text}
\usetikzlibrary{shapes.misc}
\usetikzlibrary{shapes.geometric,arrows}
\usepackage{colortbl}

\newcommand{\system}[1]{\texttt{#1}\xspace}

\newcommand{\secref}[1]{Section~\ref{#1}\xspace}
\newcommand{\tabref}[2][]{Table#1~\ref{#2}\xspace}
\newcommand{\figref}[2][]{Figure#1~\ref{#2}\xspace}

\newcommand{\equref}[1]{Equation~(\ref{#1})\xspace}

\newcommand{\bb}[1]{\mathbb{#1}}
\newcommand{\R}{\bb{R}}
\newcommand{\mat}[2][]{\boldsymbol{#2}_{#1}}

\renewcommand{\vec}[2][]{\boldsymbol{#2}^{#1}}

\newcommand{\softmax}{\textrm{softmax}}

\newcommand{\T}{\mathstrut\scriptscriptstyle\top}

\newcommand{\memnn}{\system{MemN2N}}
\newcommand{\gmemnn}{\system{GMemN2N}}

\newcommand{\RNum}[1]{\uppercase\expandafter{\romannumeral #1\relax}}

\title{Non-Markovian Control \\ with Gated End-to-End Memory Policy Networks}
\date{}

\author{
 Julien Perez\\
  Xerox Research Center Europe\\ 
 Grenoble, France\\
  \texttt{julien.perez@xrce.xerox.com} \\
  \And
 Tomi Silander\\
  Xerox Research Center Europe\\ 
 Grenoble, France\\
  \texttt{tomi.silander@xrce.xerox.com} \\
}

\begin{document}

\maketitle

\begin{abstract}
Partially observable environments present an important open challenge in the domain of sequential control learning with delayed rewards. Despite numerous attempts during the two last decades, the majority of reinforcement learning algorithms and associated approximate models, applied to this context, still assume Markovian state transitions. In this paper, we explore the use of a recently proposed attention-based model, the Gated End-to-End Memory Network, for sequential control. We call the resulting model the Gated End-to-End Memory Policy Network. More precisely, we use a model-free value-based algorithm to learn policies for partially observed domains using this memory-enhanced neural network. This model is end-to-end learnable and it features unbounded memory. Indeed, because of its attention mechanism and associated non-parametric memory, the proposed model allows us to define an attention mechanism over the observation stream unlike recurrent models. We show encouraging results that illustrate the capability of our attention-based model in the context of the continuous-state non-stationary control problem of stock trading. We also present an OpenAI Gym environment for simulated stock exchange and explain its relevance as a benchmark for the field of non-Markovian decision process learning.
\end{abstract}

\section{Introduction}
\label{sec:intro}

Reinforcement learning (RL) methods in realistic environments typically need to deal with incomplete and noisy state information resulting from partial observability as formalized by Partially Observable Markov Decision Processes (POMDPs) \cite{sondi1971}. In addition, they often need to deal with non-Markovian problems where there are significant dependencies on earlier states. Both POMDPs and non-Markovian problems largely defy traditional fully parametric value function or policy based approaches and currently require handcrafted state estimators based on accurate knowledge of the system. In this context, the use of neural networks used a value function or policy over a reinforcement learning paradigm for solving continuous control problems has a long history. Several recent papers successfully apply model-free, direct policy search methods to the problem of learning neural network control policies for challenging continuous domains with many degrees of freedom \cite{BalduzziG15, HeessWSLET15, LevineFDA15}. However, all of this work still assumes a fully observed state.

A naive alternative to using memory is learning reactive stochastic policies \cite{SinghJJ94} which simply map observations to probabilities of actions. The underlying assumption is that state-information does not play a crucial role during most parts of the problem and that using random actions can prevent the policy from getting stuck in an endless loop for ambiguous observations. Often, this strategy is far from optimal and algorithms that use some form of memory remain necessary. In summary, when a perfect model and a precisely estimated state can not be assumed, an optimal policy is likely to be memory-based. However, works on policy gradient methods with memory have been rare so far, and largely limited to finite-state controllers \cite{aber03, mel99}. More recently, some work has been proposed for a partially observable instance of the ATARI 2600 framework \cite{BellemareNVB13,HausknechtS15} but that work makes no attempt to provide an attention mechanism over an accumulative memory.Rather, those authors suggest to use a fixed size memory model, Long Short Term Memory (LSTM), for control learning.

In this paper, we extend the above LSTM approaches to more sophisticated policy representations capable of representing an observed state using a memory enhanced architecture called Gated End-to-End Memory Policy Network. With this model, policy gradient type of algorithm can effectively learn policies for POMDPs using an unbounded memory by leveraging an attention mechanism over the past observations. As a result, policy updates can depend on any event in the history. We show that our method outperforms other RL methods on a proposed benchmark task: continuous control in a non-Markovian trading environments. 

The paper is organized as follows: \secref{sec:bg} formulates POMDPs and discusses the use of reinforcement learning for POMDPs. \secref{sec:model} proposes the usage of Gated End-to-End Memory Networks for memory-enhanced reinforcement learning. In this section, a derivation of the model as policy network is presented. Then, \secref{sec:env} describes the trading and optimized execution tasks chosen for evaluation purposes. The pertinence of such environment, developed using the OpenAi Gym framework is discussed. Finally, \secref{sec:experiments} presents results the two task using $8$ real indices.

\section{Background}
\label{sec:bg}

\subsection{Markov Decision Process and Reinforcement Learning}

In the standard paradigm of Reinforcement Learning, an agent interacts with an environment $\mathcal{E}$ during a potentially infinite number of discrete time steps. At each time step $t$, the agent observes a state $s_t \in \mathcal{S}$ and chooses an action $a_t$ from some set of admissible actions $\mathcal{A}$ by using its policy $\pi$, where $\pi$ is a function from states $s_t$ to actions $a_t$. As a result, the agent observes the next state $s_{t+1}$ and receives a scalar reward $r_t$. The process continues until the agent reaches a terminal state. We define as return $R_t = \sum_{k=0}^{\infty} \gamma^k r_{t+k}$, the total accumulated return from time step $t$ with discount factor $\gamma \in (0,1]$. The goal of the agent is to maximize the expected return from each state $s_t$. The action value $Q^{\pi}(s,a) = \mathbb{E}\left[R_t|s_t=s, a\right]$ is the expected return for selecting action $a$ in state $s$ and following policy $\pi$. The optimal value function $Q^*(s,a) = \max_{\pi} Q^{\pi}(s,a)$ gives the maximum action value for state $s$ and action $a$ achievable by any policy. Similarly, the value of state $s$ under policy $\pi$ is defined as $V^{\pi}(s) = \mathbb{E}\left[R_t|s_t=s\right]$ and is simply the expected return for following policy $\pi$ from state $s$. In value-based model-free reinforcement learning methods, the action value function is often modeled using a function approximator, such as a neural network. Let $Q(s,a;\theta)$ be an approximate action-value function with parameters $\theta$. The updates to $\theta$ can be defined by a variety of reinforcement learning algorithms. A well known example of such an algorithm is Q-learning, which aims to directly approximate the optimal action value function: $Q^*(s,a)\approx Q(s,a;\theta)$. In one-step Q-learning, the parameters $\theta$ of the action value function $Q(s,a;\theta)$ are learned by iteratively minimizing a sequence of loss functions, where the $i$th loss function defined as$ L_i(\theta_i) = \mathbb{E}\left(r + \gamma \max_{a'}Q(s', a';\theta_{i-1}) - Q(s,a;\theta_i) \right)^2$ where $s'$ is the state encountered after state $s$. This standard formulation of the problem is called a Markov Decision Process. It assumes that the environment is Markovian, which means the transition to a state $s_{t+1}$ is only conditioned by the $\{s_t,a_t\}$ pair.

\subsection{Partially Observable Markov Decision Process}

Formally, a POMDP is described as a 6-tuple $(\mathcal{S}, \mathcal{A}, \mathcal{P}, \mathcal{R}, \omega, \mathcal{Z})$, where $\mathcal{S}$, $\mathcal{A}$, $\mathcal{P}$, and $\mathcal{R}$ are, respectively, the states, actions, transition function $P(S_{t+1} | S_t, A_t)$, and reward function $R : S \times A \rightarrow \mathbb{R}$ of a Markov Decision Process (MDP). In addition, the agent has no longer access to the true system state but receives an observation instead. This observation is generated from the underlying system state according to the probability distribution $z \sim \mathcal{Z}(s)= P(z_t | s_t)$. The goal of the agent is to infer a policy $\pi : \mathcal{Z}_{1:t} \rightarrow A_t$ in order to maximize cumulative reward. The formalism of POMDP well captures the dynamics of many real world environments by explicitly acknowledging that the perception received by the agent offers only a partial glimpse of the underlying state. In realistic world environments it is not reasnoable to assume that the full state of the system can be provided to the agent or even determined. Consequently, the Markov property rarely holds in such observed environments.

The resolution of partial observability, also called {\it perceptual aliasing} \cite{wblpate91}, is non-trivial and existing methods can roughly be divided into two classes. The first class of approaches explicitly maintain a belief state that corresponds to the distribution over the world states given the previous observations. Assuming a model-free hypothesis, i.e. no assumption taken over the transition and reward functions, a policy can be derived from this state estimation using reinforcement learning methods like value based methods, e.g. Q-Learning or policy based methods, e.g. policy-gradient approaches. Two major disadvantages can be mentioned: The first is the need for a model of $\mathcal{Z}(s)$ to support the state inference task. The second is the computational cost that is typically associated with the update of this belief state \cite{kaebling98, ShaniPK13}. The second class of approaches learn to form and use memories based on interactions with the environment. These methods are challenging since it is a priori unknown which features of the observations will be relevant later, and associations may have to be formed over many steps. Here, having a differentiable mechanism to learn such dependencies from experience becomes desirable. For this reason, most model free approaches tend to assume full observability. In practice, partial observability is often solved by hand-crafting a sufficient state representation from observations. As an example, in video-games, one can estimate velocity from consecutive frames \cite{mnih2015, DuanCHSA16}.

\subsection{Deep Recurrent Q-Learning}

As mentioned before, the first attempts of Deep Q-Network, experimented on ATARI 2600 video-games, had no explicit mechanism for inferring the underlying state sequence of the POMDP, thus being effective only when a contiguous series of past observations reflect of the underlying system states \cite{mnih2015}. In the general case, learning a Q-function : $S \times A \rightarrow R$ from a fixed observation window can be arbitrarily bad since $Q(z_{t-k:t}, a_t | \theta) \neq Q(s_t, a_t | \theta)$, where $k$ is fixed. More recently, Deep Recurrent Q-Learning has been proposed \cite{HausknechtS15}. This method uses a recurrent network, namely an Long Short Term Memory (LSTM) \cite{HochreiterS97}, to add a memorization capability to the previously proposed model. A drawback of this proposal comes from the necessity of selecting a priori, or using cross-validation, the dimension of the hidden and context vectors of the recurrent model which determine the memorization capacity. Another point of discussion might concern the experimental setting used in this last work. The authors propose to develop an artificial "Flickered" version of the Atari 2600 platform in order to mask parts or the entire current frame at a given period in order to force the model to memorize. In such a way, the performance of the model on an environment that has been transformed to a non-Markovian one can be measured. Finally, \cite{OhCSL16} is the closest reference to our work. The authors experiment the use of an off-the-shelf memory network as policy for the task of exploration and path-finding in a virtual 3D environment.

In this paper, we make two propositions. First, we investigate the use of a gated attention mechanism coupled with a deep recurrent Q-Network. We suggest that such mechanism may allow the Q-network to better estimate the underlying system state, narrowing the gap between $Q(z_{t-k:t}, a_t | \theta)$ and $Q(s_t, a_t | \theta)$. Indeed, in the following sections we will show that attention enhanced deep Q-networks can better approximate actual Q-values from sequences of observations, leading to better policies in partially observed environments. As a second contribution, we present a simple simulated environment of stock trading for evaluating our proposed model. We compare it to fully connected neural networks and LSTM in the tasks of stock exchange and a simplified but realistic task of optimized execution that we will now briefly present.

\subsection{Algorithmic Trading}

The field of algorithmic trading regroups a large family of methods that have been proposed to perform autonomous decision models over the global financial market. The discipline can be roughly decomposed into two categories. On the first hand, predictive methods with deterministic policies consist in learning indicators used as support for a deterministic, or stochastic but stationary, decision schema \cite{Levin95, ZimmermannNG01}. These methods consist in learning actionable patterns used to trigger buying or selling actions based on the history of a identified set of trading signals or external macro-economical informations. On the other hand, policy learning has been investigated as a way to learn a investement and portfolio management policy directly from the stock market history and also macro-economical events \cite{Neuneier95, Neuneier97, MoodyS98}. More recently, the task of optimized execution has also been studied \cite{NevmyvakaFK06}. In this context, the action space is reduced to just either selling or buying. Indeed, the actual policy been determined by a independent system, the optimized execution algorithm is in charge of applying an order to the market while leveraging on the constant fluctuation of the share prices in order to maximize the profitability of a chosen operation. In the context of this paper, we do not have the ambition to challenge highly a priori knowledge enriched and partially handcrafted portfolio management policies that are currently implemented in the real market place. However, we believe this execution context can be a novel and fruitful environment of experiment for conducting research on non-Markovian decision policy learning.

\section{Attention-based Deep Reinforcement Learning}
\label{sec:model}

\subsection{Attention models for non-Markovian reinforcement learning}

In a model-free approach, the non-Markovian observation state transitions require the decision model to store observations resulting from the interaction with the environment in order to gather sufficient information to support decision. Recently, recurrent models like LSTM have been investigated to incorporate such a memorization capability into the decision model for direct policy or value function learning \cite{HausknechtS15, LampleC16}. A drawback of such an approach is the necessity of defining, as a hyper-parameter, the dimension of the hidden state vector of the network that limits the memory of the model. Furthermore, such a recurrent model does not explicitly learn to focus its attention to different parts of a growing memory when long temporal dependencies occur in observation space.

As an alternative, attention-based models have already provided an encouraging alternative on several sequential decision tasks with immediate reward maximization like natural language translation \cite{BahdanauCB14} or end-to-end dialog systems. In the former domain, two types of approaches have been investigated. The so-called sequence-to sequence model aims at memorizing the overall source sentence before deciding the target sentence words sequentially \cite{SutskeverVL14}. The attention-based model aims at iteratively constructing a representation of an unbounded memory conditioned by the current state of the target sentence word generator \cite{BahdanauCB15, PerezL16}. Motivated by the recent empirical success of the latter method, we further investigate such an approach based on the recently proposed Gated Memory Network model.

As depicted in the next section, originally this line of research focused on text-based applications like natural language understanding, dialog management and machine reading. So, we propose to adapt and extend the use of such a model to policy learning.

\subsection{Gated End-to-End Memory Policy Networks}

The End-to-End Memory Network architecture (\memnn) \cite{Sukhbaatar+:2015}, consists of two main components: supporting memories and final answer prediction. Supporting memories are in turn comprised of a set of input and output memory representations with memory cells. The input and output memory cells, denoted by $\vec{m}_{i}$ and $\vec{c}_{i}$, are obtained by transforming the input observations $x_{1},\ldots,x_{n}$ using two embedding matrices $\mat{A}$ and $\mat{C}$, both of size $d \times d_o$ where $d$ is the embedding size and $d_o$ the dimension of the observations gathered from the environment, such that $\vec{m}_{i} = \mat{A}\Phi(x_{i})$ and $\vec{c}_{i} = \mat{C}\Phi(x_{i})$ where $\Phi(\cdot)$ is a function that maps the input into a real-valued space of dimension $d_o$. Similarly, in the original \memnn model, a question $q$ is encoded using another embedding matrix $\mat{B} \in \R^{d \times d_q}$, resulting in a question embedding $\vec{u} = \mat{B}\Phi(q)$. The input memories $\{\vec{m}_{i}\}$, together with the embedding of the question $\vec{u}$, are utilized to determine the relevance of each of the observations in the context yielding a vector of attention weights $p_{i} = \softmax(\vec[\T]{u}\vec{m}_{i})$ where $\softmax(a_{i}) = \dfrac{e^{a_{i}}}{\sum_{j \in [1, n]}e^{a_{j}}}$. Subsequently, the response $\vec{o}$ from the output memory is constructed by the weighted sum: 
\begin{equation}
\label{equ:compo}
\vec{o} = \sum_{i}p_{i}\vec{c}_{i}
\end{equation}

For more difficult tasks requiring multiple supporting memories, the model can be extended to include more than one set of input/output memories by stacking a number of memory layers. In this setting, each memory layer is named a hop and the $(k+1)^{\textrm{th}}$ hop takes as input the output of the $k^{\textrm{th}}$ hop:
\begin{equation}
\label{equ:memnn}
\vec[k+1]{u} = \vec[k]{o} + \vec[k]{u}
\end{equation}

\begin{figure*}[tb]
%\vspace{-0.5cm}
\begin{center}
\resizebox{.8\textwidth}{!}{
\begin{tikzpicture}

% Input stories
\node[draw,thick,align=center,minimum height=1.2cm,minimum width=2.8cm,rounded corners] at (7.0,0.5) (stories) {};
%\node[draw,thick,align=center,minimum height=2.0cm,minimum width=2.8cm,rounded corners] at (14.5,1.35) (stories) {};
\node[anchor=north west,shift={(0mm,2.5mm)}] (stories_xi) at (stories.west){$\{z_i\}$};
\node[draw,align=center,minimum height=0.8cm,minimum width=0.3cm] [right=0.1of stories_xi,shift={(0mm,0.3mm)}] (stories_sent1) {};
\node[draw,align=center,minimum height=0.8cm,minimum width=0.3cm] [right=0.1of stories_sent1] (stories_sent2) {};
\node[draw,align=center,minimum height=0.8cm,minimum width=0.3cm] [right=0.1of stories_sent2] (stories_sent3) {};
\node[draw,align=center,minimum height=0.8cm,minimum width=0.3cm] [right=0.1of stories_sent3] (stories_sent4) {};
\node[anchor=north west,shift={(-45mm,2mm)}] (sentences_label) at (stories.west){Observations / Stock signals};
\node[anchor=north west,shift={(-38mm,-1.2mm)}] (sentences_label) at (stories.west){$\{z_{t=1},\ldots,z_{t=T-1}\}$};
%\node[anchor=north west,shift={(-9mm,0mm)}] (sentences_label) at (stories.south){Sentences};

% Question q
\node[draw,align=center,minimum height=1.5cm,minimum width=0.3cm] at (0.7,5.5) (question) {};
\node[anchor=north west,shift={(0mm,0mm)},rotate=-90] (question_label) at (question.west){Agent specific variables $v_{k=t}$};

% Memory block 1
\node[draw,align=center,minimum height=2.5cm,minimum width=0.5cm,fill=cyan] at (2.0,3.2) (A1) {};
\node[draw,align=center,minimum height=2.5cm,minimum width=0.5cm,fill=orange] [right=0.2of A1] (C1) {};

% Tx
\node[draw,thick,align=center,circle,minimum size=0.8cm] [right=1.1 of question] (Tx1) {$\textrm{T}^{1}$};

% Element-wise product
\node[draw,thick,align=center,circle,minimum size=0.8cm] [right=0.6 of Tx1,shift={(0mm,15mm)}] (Prod11) {$\odot$};
\node[draw,thick,align=center,circle,minimum size=0.8cm] [right=0.6 of Tx1,shift={(0mm,-15mm)}] (Prod12) {$\odot$};

% Element-wise sum
\node[draw,thick,align=center,circle,minimum size=0.8cm] [right=2.0 of Tx1] (Sum1) {$\Sigma$};

% Path
\draw [rounded corners,thick,->,>=stealth] (question) -- ($(question.east) + (0.5,0.0)$) node [midway,above] (B_label) {$\mat{B}$} |- (A1);
\draw [rounded corners,thick,->,>=stealth] (stories) |- ($(A1.south) + (0.0,-0.6)$) -| (A1.south) node [left,shift={(0mm,-3mm)}] (A1_label) {$\mat[1]{A}$};
\draw [rounded corners,thick,->,>=stealth] (stories) |- ($(C1.south) + (0.0,-0.6)$) -| (C1.south) node [right,shift={(0mm,-3mm)}] (C1_label) {$\mat[1]{C}$};

\draw [rounded corners,thick,->,>=stealth] (question) -- (Tx1) node [above,shift={(-6mm,0mm)}] (u1_label) {$\vec[1]{u}$};
\draw [rounded corners,thick,->,>=stealth] (question) -- ($(question.east) + (0.5,0.0)$) |- (Prod11) node [above,shift={(-6mm,0mm)}] (u1_label) {$\vec[1]{u}$};

\draw [rounded corners,thick,->,>=stealth] (Tx1) -- ($(Tx1.east) + (0.8,0.0)$) -| (Prod11.south) node [left,shift={(0mm,-2mm)}] () {$1 - \textrm{T}^{1}(\vec[1]{u})$};
\draw [rounded corners,thick,->,>=stealth] (Tx1) -- ($(Tx1.east) + (0.8,0.0)$) -| (Prod12.north) node [left,shift={(0mm,4mm)}] () {$\textrm{T}^{1}(\vec[1]{u})$};

\draw [rounded corners,thick,->,>=stealth] (C1.east) -- ($(C1.east) + (0.3,0.0)$)  node [above,shift={(0mm,0mm)}] () {$\vec[1]{o}$} -| (Prod12.south);

\draw [rounded corners,thick,->,>=stealth] (Prod11.east) -- ($(Prod11.east) + (0.3,0.0)$) -| (Sum1.north);
\draw [rounded corners,thick,->,>=stealth] (Prod12.east) -- ($(Prod12.east) + (0.3,0.0)$) -| (Sum1.south);

% Memory block 2
\node[draw,align=center,minimum height=2.5cm,minimum width=0.5cm,fill=cyan] [right=3.4 of C1] (A2) {};
\node[draw,align=center,minimum height=2.5cm,minimum width=0.5cm,fill=orange] [right=0.2of A2] (C2) {};

% Tx
\node[draw,thick,align=center,circle,minimum size=0.8cm] [right=0.9 of Sum1] (Tx2) {$\textrm{T}^{2}$};

% Element-wise product
\node[draw,thick,align=center,circle,minimum size=0.8cm] [right=0.6 of Tx2,shift={(0mm,15mm)}] (Prod21) {$\odot$};
\node[draw,thick,align=center,circle,minimum size=0.8cm] [right=0.6 of Tx2,shift={(0mm,-15mm)}] (Prod22) {$\odot$};

% Element-wise sum
\node[draw,thick,align=center,circle,minimum size=0.8cm] [right=2.0 of Tx2] (Sum2) {$\Sigma$};

% Path
\draw [rounded corners,thick,->,>=stealth] (Sum1) -- ($(Sum1.east) + (0.4,0.0)$) |- (A2);
\draw [rounded corners,thick,->,>=stealth] (stories) |- ($(A2.south) + (0.0,-0.6)$) -| (A2.south) node [left,shift={(0mm,-3mm)}] (A2_label) {$\mat[2]{A}$};
\draw [rounded corners,thick,->,>=stealth] (stories) |- ($(C2.south) + (0.0,-0.6)$) -| (C2.south) node [right,shift={(0mm,-3mm)}] (C2_label) {$\mat[2]{C}$};

\draw [rounded corners,thick,->,>=stealth] (Sum1) -- (Tx2) node [above,shift={(-6mm,0mm)}] (u2_label) {$\vec[2]{u}$};
\draw [rounded corners,thick,->,>=stealth] (Sum1) -- ($(Sum1.east) + (0.4,0.0)$) |- (Prod21) node [above,shift={(-6mm,0mm)}] (u2_label) {$\vec[2]{u}$};

\draw [rounded corners,thick,->,>=stealth] (Tx2) -- ($(Tx2.east) + (0.8,0.0)$) -| (Prod21.south) node [left,shift={(0mm,-2mm)}] () {$1 - \textrm{T}^{2}(\vec[2]{u})$};
\draw [rounded corners,thick,->,>=stealth] (Tx2) -- ($(Tx2.east) + (0.8,0.0)$) -| (Prod22.north) node [left,shift={(0mm,4mm)}] () {$\textrm{T}^{2}(\vec[2]{u})$};

\draw [rounded corners,thick,->,>=stealth] (C2.east) -- ($(C2.east) + (0.3,0.0)$)  node [above,shift={(0mm,0mm)}] () {$\vec[2]{o}$} -| (Prod22.south);

\draw [rounded corners,thick,->,>=stealth] (Prod21.east) -- ($(Prod21.east) + (0.3,0.0)$) -| (Sum2.north);
\draw [rounded corners,thick,->,>=stealth] (Prod22.east) -- ($(Prod22.east) + (0.3,0.0)$) -| (Sum2.south);

% Memory block 3
\node[draw,align=center,minimum height=2.5cm,minimum width=0.5cm,fill=cyan] [right=3.4 of C2] (A3) {};
\node[draw,align=center,minimum height=2.5cm,minimum width=0.5cm,fill=orange] [right=0.2of A3] (C3) {};

% Tx
\node[draw,thick,align=center,circle,minimum size=0.8cm] [right=0.9 of Sum2] (Tx3) {$\textrm{T}^{3}$};

% Element-wise product
\node[draw,thick,align=center,circle,minimum size=0.8cm] [right=0.6 of Tx3,shift={(0mm,15mm)}] (Prod31) {$\odot$};
\node[draw,thick,align=center,circle,minimum size=0.8cm] [right=0.6 of Tx3,shift={(0mm,-15mm)}] (Prod32) {$\odot$};

% Element-wise sum
\node[draw,thick,align=center,circle,minimum size=0.8cm] [right=2.0 of Tx3] (Sum3) {$\Sigma$};

% Path
\draw [rounded corners,thick,->,>=stealth] (Sum2) -- ($(Sum2.east) + (0.4,0.0)$) |- (A3);
\draw [rounded corners,thick,->,>=stealth] (stories) |- ($(A3.south) + (0.0,-0.6)$) -| (A3.south) node [left,shift={(0mm,-3mm)}] (A3_label) {$\mat[3]{A}$};
\draw [rounded corners,thick,->,>=stealth] (stories) |- ($(C3.south) + (0.0,-0.6)$) -| (C3.south) node [right,shift={(0mm,-3mm)}] (C3_label) {$\mat[3]{C}$};

\draw [rounded corners,thick,->,>=stealth] (Sum2) -- (Tx3) node [above,shift={(-6mm,0mm)}] (u3_label) {$\vec[3]{u}$};
\draw [rounded corners,thick,->,>=stealth] (Sum2) -- ($(Sum2.east) + (0.4,0.0)$) |- (Prod31) node [above,shift={(-6mm,0mm)}] (u3_label) {$\vec[3]{u}$};

\draw [rounded corners,thick,->,>=stealth] (Tx3) -- ($(Tx3.east) + (0.8,0.0)$) -| (Prod31.south) node [left,shift={(0mm,-2mm)}] () {$1 - \textrm{T}^{3}(\vec[3]{u})$};
\draw [rounded corners,thick,->,>=stealth] (Tx3) -- ($(Tx3.east) + (0.8,0.0)$) -| (Prod32.north) node [left,shift={(0mm,4mm)}] () {$\textrm{T}^{3}(\vec[3]{u})$};

\draw [rounded corners,thick,->,>=stealth] (C3.east) -- ($(C3.east) + (0.3,0.0)$)  node [above,shift={(0mm,0mm)}] () {$\vec[3]{o}$} -| (Prod32.south);

\draw [rounded corners,thick,->,>=stealth] (Prod31.east) -- ($(Prod31.east) + (0.3,0.0)$) -| (Sum3.north);
\draw [rounded corners,thick,->,>=stealth] (Prod32.east) -- ($(Prod32.east) + (0.3,0.0)$) -| (Sum3.south);

% Final prediction
\node[draw,thick,align=center,minimum width=0.8cm] [right=0.5 of Sum3] (W) {$\textbf{\textit{W}}$};

\node[draw,thick,align=center,minimum height=1.5cm] [below=0.5 of W] (a) {$\hat{\textbf{\textit{Q(z,a)}}}$};
\node[anchor=north west,shift={(-8mm,-9mm)},rotate=90] (answer_label) at (a.west){Predicted};
\node[anchor=north west,shift={(-5mm,-9mm)},rotate=90] (answer_label) at (a.west){Actions};

\draw [rounded corners,thick,->,>=stealth] (Sum3.east) -- (W.west);
\draw [rounded corners,thick,->,>=stealth] (W.south) -- (a.north);

\end{tikzpicture}
}
\end{center}
%\vspace{-1.0cm}
\caption{\label{fig:resmemn2n}Illustration of the proposed \gmemnn model with $3$ hops.}
\end{figure*}
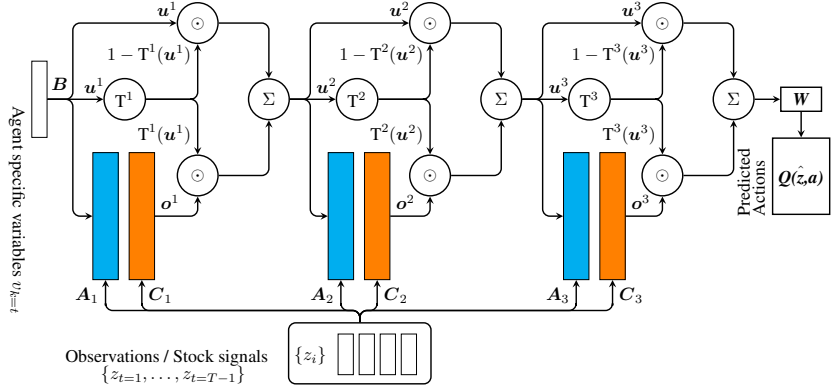

Lastly, the final step, the prediction of the answer to the question $q$, is performed by $\hat{\vec{a}} = \softmax(\mat{W}(\vec[K]{o} + \vec[K]{u}))$ where $\hat{\vec{a}}$ is the predicted answer distribution, $\mat{W} \in \R^{|V| \times d}$ is a parameter matrix for the model to learn and $K$ the total number of hops. As suggested in \cite{PerezL16}, \equref{equ:memnn} can be considered as a form of residual with $\vec[k]{o}$ working as the residual function and $\vec[k]{u}$ the shortcut connection. However, as discussed in \cite{Srivastava+:2015}, in contrast to the hard-wired skip connection in Residual Networks, one of the advantages of Highway Networks is the adaptive gating mechanism, capable of learning to dynamically control the information flow based on the current input. Therefore, we adopt the idea of the adaptive gating mechanism of Highway Networks and integrate it into \memnn. The resulting model, named \textit{Gated End-to-End Memory Networks} (\gmemnn) \cite{PerezL16} and illustrated in \figref{fig:resmemn2n}, is capable of dynamically conditioning the memory reading operation on the controller state $\vec[k]{u}$ at each hop. Concretely, we reformulate \equref{equ:memnn} into:
\begin{align}
\textrm{T}^{k}(\vec[k]{u}) &= \sigma(\mat[T]{W}^{k}\vec[k]{u} + \vec[k]{b}_T) \\
\vec[k+1]{u}  &= \vec[k]{o}\odot\textrm{T}^{k}(\vec[k]{u}) + \vec[k]{u}\odot(1 - \textrm{T}^{k}(\vec[k]{u}))
\end{align}

where $\mat[T]{W}^{k}$ and $\vec[k]{b}$ are the hop-specific parameter matrix and bias term for the $k^{\textrm{th}}$ hop and $\textrm{T}^{k}(x)$ the transform gate for the $k^{\textrm{th}}$ hop and $\sigma$ a sigmoidal activation function.

In the case of policy learning, the memory cells are filled with past observations collected from past interactions with the environment, and the question input will carry current state information that are relevant to the agent and independent from the environment observations. In the context of stock trading and optimized execution, the memory blocks will carry the past values of the traded signal and the question block will carry the current budget and portfolio composition of the agent. Finally, assuming a discrete action set, the output of the model, the answer, will be the expected reward associated to each eligible action. \figref{fig:resmemn2n} summarizes the elements of this Gated End-to-End Memory Policy Network.

\subsection{Neural Temporal Encoding}

A limitation of Memory Networks compared to other types of attention-based models, like those applied to machine translation, is the necessity to encode temporal information into the memory blocks. Indeed, because of the commutative nature of Equation \ref{equ:compo}, any information regarding the order of the observations embedded in the memory blocks has to be encoded beforehand. In Dynamic Memory Network \cite{KumarIOIBGZPS16}, the hidden state of an LSTM is used to encode the values put into the memory blocks before computing the attention values over them. In the original end-to-end memory network~\cite{SukhbaatarSWF15}, the encoding is done using a deterministic function that transforms the sequence of word embeddings of each sentence before putting them into the memory blocks. We propose to embed the signal using a denoising and predictive neural auto-encoder. More specifically, the single hidden layer of the perceptron reconstructing the noisy input of the time frame is placed into the memory blocks. On one hand, the model is only a denoiser, but on the other hand, by adding output to the model, the neural network can predict future windows regarding the encoded time frame. This approach is related to the context-dependent word vectorization \cite{mikolov13,PenningtonSM14}.

\subsection{Policy gradient over Gated End-to-End Memory Policy Networks}

We consider policies represented as gated memory networks. The model builds a vector, i.e the controller state $u$, representing its latent state from the multiple attention-based readings of its memory blocks where the environment observations have been stored. The latent state begins with a fixed state $u_0$. At each time-step $t = 1,2, \ldots, n$, the network takes as an input a series of observations, and computes its internal state according to a differentiable function $F(z_{1:n} | \theta_f) = u_t$ and outputs a distribution over actions $a_t$ according to a differentiable function $G(u_t|\theta_g)=a_t$ where $\theta=(\theta_f, \theta_g)$. $\pi^\theta(a_t|o_{1:t})$ denotes the output of the memory network at time-step $t$. Past work has defined a principled method for updating the parameters $\theta$ of the policy $\pi^\theta$ through reinforcement learning \cite{Williams92, PetersS06} using stochastic gradient descent:$\bigtriangleup\theta_d = \sum^{T-1}_{t=0}\bigtriangledown_\theta \mbox{log} \pi^\theta(a_t|z_{1:t})G_t$. While this update is unbiased, in practice it is known to suffer high variance and low converge rate. It has been shown \cite{Williams92} that this update can be rewritten as $\bigtriangleup\theta_d = \sum^{T-1}_{t=0}\bigtriangledown_\theta \mbox{log} \pi^\theta(a_t|z_{1:t})(G_t - b)$, where $b$ is a baseline, which can be an arbitrary function of states visited during an episode. Using this general framework of policy-gradient learning via Gated Memory Network, we define our control model using the approach that have been described in the context of language modeling \cite{SukhbaatarSWF15}. In this application, a constant is defined as $q$ and the network produces as an output a distribution over the vocabulary. This kind of approach can be put in parallel with the control model of Deep Q-Learning proposed in \cite{mnihdqn2015} where a Convolutional Neural Network takes as input a contiguous sliding window of video game screens and output the Q-values associated to a finite set of eligible actions. Finally, because of its stability in learning parametric policies, we use Asynchronous Deep Q-Learning as reinforcement learning algorithm which is described in Algorithm \ref{alg:targetq} as proposed in \cite{MnihBMGLHSK16}.

\section{Trading and Optimized Execution}
\label{sec:env}

\subsection{Trading Environment}

As an evaluation environment, we developed a simplified portfolio management platform. Following the settings proposed in \cite{MoodyS98}, the decision space of trading consists in a set of three discrete actions $A \in \{Buy, Hold, Sell\}$ assuming a fixed amount of stock exchanged for each action. The observation space $\mathcal{Z} \in \mathbb R^k$ is the current value of the $k$ stocks considered for trading. For each transaction, a fixed transaction cost is associated. In a more realistic setting, the transaction cost is likely to be a function of the type and the amount of stocks involved at each decision step. In our experiments, we only consider the task of speculative trading which means that the reward, measured as the increase of budget at a given time step is the result of the evolution of the market shares. In a more realistic settings, dividends, which are the part of the companies benefice distributed to share holders, should also be considered as a potential source of income, especially in a multi-year scale and multi-stocks management settings.

A second task that as been studied in the litterature is the optimized execution setting. It consists of either selling or buying a given amount of stock in a fixed amount of time as described in \cite{NevmyvakaFK06}. For the optimized buying case, the goal consists in buying the desired amount of stock at the cheapest price over a given period of time. For the optimized selling case, the goal consists in following an acquisition strategy that allows us to sell at the higher possible price during the given period. Our simulation platform has been developed as a OpenAI Gym \cite{BrockmanCPSSTZ16} environment and is planned to be published as an open-source package. Our purpose is to encourage the research community of non-Markovian reinforcement learning to use such a framework as a reusable experimental testbed.

\subsection{Trading signals and Attention Based Controllers}

During our experiments, indices have been studied as trading signals. The daily opening prices of a set of real indices have been chosen. However an other advantage of using stock exchange as a test-bed for non-Markovian control is the possibility to also generate such synthetic series. In comparison to other virtual environment, like First Person Shooter \cite{KempkaWRTJ16} or Atari 2600 \cite{BellemareNVB13}, the control of the required memory capacity to perform profitable control can be defined by estimating the Markovian order of the series. Indeed, in the context of games, the memory capacity can hardly be related from the partially observable maze or first person shooter as a function of the size of the maze. However, in the case of trading, the memory capacity requirement can be defined as the order of the time series. For our experiments, we choose to focus on 8 real indices taken from the main market places in US, Europe and Asia.

Our Gated End-to-End Memory Policy Network takes as input the past observations of the traded series. At each time step, It computes the expected reward of each eligible actions. The model is optimized through policy gradient, prioritized experience replay \cite{SchaulQAS15} and double Q-learning in order to cope with inherent instability of such learning process. Beyond the stability and convergence rate compared to Q-Learning, such model allows one to implement a Boltzmann type of policy over the reward expectation using one forward pass of the model.

\section{Experiments}
\label{sec:experiments}

\subsection{Training Details}
\label{sec:trainingdetails}

Concernint the parameterization of our decision model. As suggested in \cite{Sukhbaatar+:2015}, $10\%$ adjacent weight tying, and temporal encoding with $10\%$ random noise is used. Learning rate $\eta$ is initially assigned a value of $0.001$ with exponential decay applied every $30$ epochs by $\eta/2$ until $100$ epochs are reached. Linear start is used in all our experiments as proposed by \cite{Sukhbaatar+:2015}. With linear start, the $\softmax$ in each memory layer is removed and re-inserted after $30$ epochs. Batch size is set to $32$ and gradients with an $\ell_2$ norm larger than $10$ are divided by a scalar to have norm $10$. All weights are initialized randomly from a Gaussian distribution with zero mean and $\sigma = 0.1$ except for the transform gate bias $\vec[k]{b}_{T}$ which we empirically set the mean to $0.2$. In all our experiments, we use the embedding size $d = 20$. As in \cite{Sukhbaatar+:2015}, since the memory-based models are sensitive to parameter initialization, we repeat each training $20$ times and choose the best system based on the performance on the validation set. The temporal neural encoders are learnt individually over each training series and used in test to preprocess observation sequences before been placed into the memory block of the policy network. The hidden layer dimension of each encoder has been set by cross-validation to $25$ and optimized using Adam \cite{Kingma+:2014}. Then, the baseline neural policy network is composed with two hidden layers of $30$ hidden units with rectified linear activation and a linear output projection. The baseline LSTM model has a hidden representation of $50$ dimensions. All the hyperparameters haven been estimated through cross-validation. Concerning the policy learning algorithm. The reward function is episodic. At the end of each episode, the agent receives a reward which is the difference between the budget at the end of the period and the initial budget. The network was trained using $200$ consecutive days of daily opening values. The training phase consists in 10000 trading episodes over these sequences of values. The training on a given series represents approximatively one hour on one core of a NVIDIA Tesla P-100 GPU. In this experiment all policies are learnt independently from one series to another. The testing phase of each trading experiment is performed using $200$ consecutive days of market. In the case of optimized trading, each testing corresponds to $100$ roll-outs. The resulting policies follows a Bolzmann distribution over the reward predicted by the policy network. Finally, the update period of the Double Q-Learning mechanism is $100$ action steps. 

\subsection{Results and discussions}

\tabref{tab:prof} computes the profitability ratio which corresponds to the number of days, over the test period, where the agent is profitable. A trading day is qualified as profitable if the difference between the corresponding current budget and the initial budget of the agent is positive. Such evaluation makes sense as a speculative strategy where maximizing the amount of positive market exit opportunities over a given period of time is excepted to be maximized. This first results confirms the utility of a control policy equipped with a memorization capability. Then, a control policy equipped with an attention mechanism as the one proposed in this work seems to be confirm.

\begin{figure}[!ht]
\small
\center
\begin{tabular} {|l|l|l|l|}
\hline
Indices & Policy Network & Profitability ratio & Resulting budget \\
\hline
&FCNN & $0.46\pm 0.037$ & $31.98\pm 0.14$\\
CAC40&LSTM & $0.49\pm 0.023$ & $37.96\pm 0.16$\\
&MemN2N & $ 0.51\pm 0.015$ & $39.01\pm 0.12$\\
&GMemN2N & ${\bf 0.53\pm 0.014}$ & ${\bf 39.97\pm 0.12}$\\
\hline
&FCNN & $0.49\pm 0.30$ & $39.97\pm 0.20$\\
GDAXI&LSTM & $0.54\pm 0.048$ & $50.47\pm 0.08$\\
&MemN2N & $0.57\pm 0.02$  & $51.1\pm 0.09$\\
&GMemN2N & ${\bf 0.59\pm 0.019}$  & ${\bf 52.32\pm 0.07}$\\
\hline
&FCNN & $0.47\pm 0.034$ & $39.97\pm 0.14$\\
JKII&LSTM & $0.48\pm 0.025$ & $42.8\pm 0.09$\\
&MemN2N & $0.50\pm 0.014$ & $43.48\pm 0.09$\\
&GMemN2N & ${\bf 0.51\pm 0.017}$ & ${\bf 46.48\pm 0.10}$\\
\hline
&FCNN & $0.44\pm 0.034$ & $31.95\pm 0.21$\\
NASDAQ100&LSTM & $0.45\pm 0.013$  & $49.01\pm 0.04$\\
&MemN2N & $0.47\pm 0.028$ & $50.79\pm 0.09$\\
&GMemN2N & ${\bf 0.49\pm 0.013}$ & ${\bf 51.21\pm 0.12}$\\
\hline
&FCNN & $0.45\pm 0.030$ & $41.9\pm 0.12$\\
NIKKEI225&LSTM & $0.55\pm 0.052$ & $47.76\pm 3.27$\\
&MemN2N & $0.57\pm 0.048$ & $47.98\pm 0.14$\\
&GMemN2N & ${\bf 0.59\pm 0.042}$ & ${\bf 49.98\pm 0.14}$\\
\hline
&FCNN & $0.49\pm 0.033$ & $49.96\pm 0.19$\\
NYSE100&LSTM & $0.43\pm 0.034$ & $50.12\pm 0.02$\\
&MemN2N & $0.53\pm 0.013$  & $51.96\pm 0.19$\\
&GMemN2N & ${\bf 0.56\pm 0.033}$  & ${\bf 53.96\pm 0.19}$\\
\hline
&FCNN & $0.38\pm 0.032$ & $31.97\pm 0.15$\\
N100&LSTM & $0.41\pm 0.065$ & $38.80\pm 0.02$\\
&MemN2N & $0.43\pm 0.044$ & $39.89\pm 2.18$\\
&GMemN2N & ${\bf 0.45\pm 0.054}$ & ${\bf 41.89\pm 2.34}$\\
\hline
&FCNN & $0.43\pm 0.023$ & $38.58\pm 1.28$\\
RUT&LSTM & $0.51\pm 0.015$ & $39.97 \pm 0.05$\\
&MemN2N & $0.53\pm 0.038$ & $45.80\pm 0.11$\\
&GMemN2N & ${\bf 0.55\pm 0.035}$ & ${\bf47.80\pm 0.15}$\\
\hline
\end{tabular}
\caption{\label{tab:prof} Profitability ratios for trading and Resulting budget in optimized execution}
\end{figure}

The evaluation of the proposed policy over an optimized selling task is also depicted. In such setting, the set of authorized actions are reduced to $\mathcal{A}=\{Hold, Sell\}$. The agent starts each episode with 50 stocks to sell in the trading period. The reward is the resulting accumulated budget at the end of the period. As for trading, the policies are evaluated of a testing series of $100$ trading days. In such settings, the proposed policy show encouraging result that confirm the benefit of an attention based mechanism of memory management for learning differentiable policies in non-Markovian environment. For all experiments, the series absolute values are max-normalized in order to accelerate gradient descent and control gradient magnitude. Finally, the necessity of a memory in such task seems to be confirmed by the inferior performance of a memory-less fully connected layer model.

%\begin{figure*}[!ht]
%
%  \includegraphics[scale=0.3]{imgs/budget/actions_N100.png}
%  \includegraphics[scale=0.3]{imgs/budget/actions_RUT.png}
%  \includegraphics[scale=0.3]{imgs/budget/actions_SP500.png}
%
%  \includegraphics[scale=0.3]{imgs/budget/actions_cac40.png}
%  \includegraphics[scale=0.3]{imgs/budget/actions_GDAXI.png}
%  \includegraphics[scale=0.3]{imgs/budget/actions_JKII.png}
%
%  \caption{\label{fig:tradeaction}Distribution of the decisions on test period of 3 months, 100 trading days on major indices.}
%\end{figure*}

%\begin{figure*}[!ht]
%
%  \includegraphics[scale=0.3]{imgs/budget/N100.png}
%  \includegraphics[scale=0.3]{imgs/budget/RUT.png}
%  \includegraphics[scale=0.3]{imgs/budget/SP500.png}
%
%  \includegraphics[scale=0.3]{imgs/budget/cac40.png}
%  \includegraphics[scale=0.3]{imgs/budget/GDAXI.png}
%  \includegraphics[scale=0.3]{imgs/budget/JKII.png}
%
%  \caption{\label{fig:tradeperf}Budget on test period of 3 months, 100 trading days on major indices.}
%\end{figure*}

\section{Conclusion and Future Work}

In this paper, we have studied the question of non-Markovian decision processes and the use of an attention-based policy network called Gated End-to-End Memory Policy Network. The task of stock exchange and optimized execution have been used as experimental testbed to illustrate the capability of the model. In addition to the proposed the model, we think such a trading environment can produce fruitful research in the domain of non-Markovian control in the future. Indeed, the settings of stock exchange and revenue maximization allow to study the behavior of policy learning algorithms and policy models with signals exhibiting different requirement of memorization. Furthermore, tasks of resource allocation and scheduling can be easily related to this formal setting. Finally, in comparison to the current results using parametric memories like Gated Rectified Units or Long Short Term Memory, we believe attention-based models, which have already demonstrated their advantages in the domain of sequence prediction in Natural Language Processing like machine translation and machine reading, can be of a first importance in the more general case of non-Markovian control. In the near future, we plan to release an open-source package of our OpenAI Gym Trade environment and the corresponding Gated Memory Policy Networks.

\bibliography{main}
\bibliographystyle{alpha}

\end{document}